\pdfoutput=1
\documentclass{article}

\usepackage[preprint]{neurips_2024}

\usepackage[utf8]{inputenc}
\usepackage[T1]{fontenc}
\usepackage{hyperref}
\usepackage{url}
\usepackage{booktabs}
\usepackage{amsfonts}
\usepackage{amsmath}
\usepackage{nicefrac}
\usepackage{microtype}
\usepackage{xcolor}
\usepackage{graphicx}
\usepackage{placeins}

\newcommand{\dmodel}{d_{\mathrm{model}}}

\title{Don't Forecast the Noise: Latent-Space Pretraining for
  Time-Series Foundation Models}

\author{%
  Jeremy Sylvain Cochoy \\
  \texttt{jeremy.cochoy@gmail.com}
}

\begin{document}

\maketitle

\begin{abstract}
  Current time-series foundation models forecast in value space,
  following the language-modelling recipe of next-token prediction.
  Because the observed series is one realisation of a stochastic
  process, a value-space loss trains the model against targets
  that carry intrinsic noise that no model can predict. We propose \emph{Latent
  Forecasting}, a training framework in which a patch embedding, an
  encoder and a causal forecaster are trained jointly to predict the
  next patch in latent space. The forecasts align with the latents of
  an EMA teacher, a contrastive loss separates the latents of
  different patches and series, and the SIGReg regulariser prevents
  the collapse of the latents. The framework is decoupled from the
  choice of encoder and forecaster architecture. A small head, trained
  on the frozen model, maps the latent forecasts back to values. We
  train and evaluate on GIFT-Eval. We demonstrate the feasibility of
  training a forecasting model in latent space only. The downstream forecast
  in value space matches the performance of a value-space model
  trained with the rest of our recipe.
  After training on about $20\%$ of the
  GIFT-Eval pretraining corpus, our model scores a GM-MASE of
  $0.93$. A Moirai-style
  value-space model of a similar size scores $0.94$.
\end{abstract}

\section{Introduction}
\label{sec:introduction}

Deep learning improved forecasting in steps:
DeepAR~\citep{salinas2020deepar} trains one probabilistic recurrent
network on many related series, N-BEATS~\citep{oreshkin2020nbeats}
forecasts with deep stacks of simple blocks, and
PatchTST~\citep{nie2023patchtst} cuts the series into patches that a
transformer reads. Time-series forecasting has converged on a
foundation-model recipe inherited from language modelling: patched
inputs, a transformer backbone, and a head that forecasts the next
patch in value space.
TimesFM~\citep{das2024timesfm} is the canonical example, and recent
successors~\citep{ansari2024chronos, goswami2024moment, woo2024moirai,
shi2024timemoe, liu2025sundial, liu2026timers1, mistral2026timestral}
extend the recipe along various axes while keeping the value-space
target. As Ilya Sutskever has
argued for large language models, next-token prediction is far less
narrow than it looks: it forces
the model to encode grammar, style, world knowledge and arithmetic,
since each is a sub-task of the same prediction problem. The
next-patch analogue should inherit this property for time series.
The choice of value space, however, is suboptimal: the
observed series is one realisation of a stochastic process, so
the targets of a value-space loss carry the intrinsic noise of
that process. No model can predict this noise,
and the training loss never reaches zero, even on processes whose
predictable component the model captures perfectly.

A second line of work shifts the prediction target from raw values to
a learned representation. The Joint Embedding Predictive
Architecture (JEPA)~\citep{assran2023ijepa}, originally introduced
for vision, encodes a context block with one network and a target
block with another, and trains a predictor to map the context latent
to the target latent. The recent time-series
adaptations~\citep{verdenius2024latpfn, ennadir2025tsjepa} apply the
same idea to univariate series. By predicting an abstract feature of
the future rather than its raw values, the model is no longer
punished for missing the noise. The siamese architecture used by
JEPA, however, is at risk of collapse by construction: the constant
function is a global minimum of its loss. Contrastive learning with
explicit negatives is the classical anti-collapse remedy, and has
been used at scale for time-series representation learning
(CPC~\citep{oord2018cpc}, TS2Vec~\citep{yue2022ts2vec}), but to our
knowledge no published time-series foundation model combines it
with a forecasting objective in latent space.

We ask whether a time-series forecasting model can be trained
directly in latent space with an explicit contrastive objective, and
how its forecasts compare with those of a value-space model. The rest
of the paper proposes a candidate architecture and evaluates it on
the GIFT-Eval benchmark, against value-space models trained with the
same data for the same number of training steps.

The contributions of the paper are the following.
\begin{itemize}
\item A training framework, \emph{Latent Forecasting}, in which
  a patch embedding, an encoder and a causal forecaster are trained
  jointly in latent space. The forecasts align with the latents of an
  EMA teacher, an InfoNCE-style objective contrasts the latents of different
  patches and series, while SIGReg prevents the collapse
  (Section~\ref{sec:method}). The framework is decoupled from the
  choice of encoder and forecaster architecture.
\item A comparison on GIFT-Eval with Moirai-style value-space models
  trained with the same data for the same number of training
  steps (Section~\ref{sec:results}), and the ablations of
  our recipe (Section~\ref{sec:ablation}).
\end{itemize}

\section{Related Work}
\label{sec:related-work}

The foundation models of Section~\ref{sec:introduction} compute their
loss on the values of the series. Two earlier lines of work move the
prediction of the future into a latent space: Contrastive Predictive
Coding and the Joint Embedding Predictive Architecture.

Contrastive Predictive Coding (CPC)~\citep{oord2018cpc} encodes each
step of a signal with a strided convolutional network, and a GRU
summarises the past latents into a context vector. For each horizon
$k$, a linear map of the context predicts the latent of the step
$t + k$. The InfoNCE loss~\citep{oord2018cpc} trains the model to
find the true future latent among negative latents, sampled from other
times and other sequences. This loss is a lower bound on the mutual
information between the context and the future, so the encoder keeps
the part of the past that predicts the future. CPC learns
representations for speech, images, text and reinforcement learning.
In speech and vision, the evaluation trains a linear classifier on the
frozen representations, for example to recognise phones or speakers.
CPC does not try to forecast. No part of the model maps a predicted
latent back to the values of the signal, and the paper reports no
forecast error. The time-series successors keep the contrastive loss
and change the positive pairs: a subseries and a shorter subseries
inside it in T-Loss~\citep{franceschi2019tloss}, temporal
neighbourhoods in TNC~\citep{tonekaboni2021tnc}, and two augmented
views of the same time step in TS2Vec~\citep{yue2022ts2vec}. These
methods target classification and anomaly detection. TS2Vec also
reports forecasts, from a ridge regression that it fits on the frozen
representations of each dataset.

The Joint Embedding Predictive Architecture
(JEPA)~\citep{assran2023ijepa} comes from vision. A context encoder
reads the visible part of an image. A target encoder reads the full
image, and its latents of the masked blocks are the targets. The
weights of the target encoder are an exponential moving average (EMA)
of the weights of the context encoder. A predictor maps the context
latents to the target latents with an $L^2$ loss. JEPA uses no
negatives. The stop-gradient on the target branch and the asymmetric
predictor prevent the collapse to a constant, which is a global
minimum of the loss. TS-JEPA~\citep{ennadir2025tsjepa} applies this
recipe to time series. It cuts each series into patches, masks
$70\%$ of them, and trains with an $L^1$ loss in latent space. It
trains one model for each dataset and evaluates it on classification
and on forecasting. In its short-term forecast experiment, the
autoregressive baseline of the paper has a lower error than TS-JEPA on
two of the three datasets.
LaT-PFN~\citep{verdenius2024latpfn} combines JEPA with a prior-data
fitted network for zero-shot forecasting in context.
LeJEPA~\citep{balestriero2025lejepa} removes the EMA teacher and the
stop-gradient. It adds SIGReg, a regulariser that pushes the
distribution of the embeddings to an isotropic Gaussian, and this
regulariser prevents the collapse.

Our model takes from each of these works. It forecasts the next patch
with a causal transformer, as TimesFM~\citep{das2024timesfm} does. Its
forecast is a latent, as in CPC and JEPA, and it must match an EMA
target, as in JEPA. A contrastive loss with negatives, as in CPC, and
the SIGReg regulariser of LeJEPA keep the latents away from collapse.
Unlike CPC and TS-JEPA, it trains one model on many datasets for
forecasting, and we score it with a forecast error on the GIFT-Eval
benchmark~\citep{aksu2024gifteval}.

\section{Latent Forecasting}
\label{sec:method}

Latent Forecasting predicts the next patch of a series in latent
space. The model has three stages: a patch embedding, an encoder and a
forecaster. A small head, trained after the model, maps the latent
forecasts back to values. The subsections below define each part, then
the loss and the head.

\subsection{Architecture}
\label{sec:architecture}

We work with univariate series: each variate of a multivariate series
is a separate series. A training window has $T = 1024$ values. We
normalise it (Section~\ref{sec:normalisation}) and cut it into
$T_p = T / W = 64$ non-overlapping patches of $W = 16$ values. Each
stage of the model writes one latent of size $\dmodel$ for each patch.
\begin{itemize}
\item The patch embedding maps patch $t$ to the embedding $e_t$
  (Section~\ref{sec:patch-embedding}).
\item The encoder, a causal transformer, maps $e_0, \ldots, e_t$ to
  the encoder latent $h_t$.
\item The forecaster, a causal transformer, maps $h_0, \ldots, h_t$ to
  $f_t$, the forecast of the latent of patch $t + 1$.
\end{itemize}

After each stage, we put the latent on the sphere of radius
$\sqrt{\dmodel}$. The encoder reads $e_t$ on the sphere, the forecaster
reads $h_t$ on the sphere, and the forecast $f_t$ is on the sphere. We
preferred a sphere of radius $\sqrt{\dmodel}$ because a vector on it has a
root mean square of $1$ for each coordinate, which is the scale of the
output of a layer normalisation. During a \emph{latent rollout}, the
forecaster also reads its own forecasts in latent space
(Sections~\ref{sec:loss} and~\ref{sec:forecast-head}), and the sphere
fixes the norm of these inputs too.

\subsection{Normalisation}
\label{sec:normalisation}

The series of a pretraining corpus have levels and amplitudes that
differ by orders of magnitude, so a foundation model normalises each
series before it reads it. A standard method is instance
normalisation, as in RevIN~\citep{kim2022revin}. It computes one mean
$\mu$ and one standard deviation $\sigma$ over the observed values of
the context, and the model reads $(x_t - \mu) / \sigma$. The forecast
$\hat{y}$ goes back to values as $\hat{y}\,\sigma + \mu$.
Moirai~\citep{woo2024moirai} uses this method, and computes $\sigma$ as
the square root of the unbiased variance plus $10^{-5}$.

Our forecaster predicts the next patch at every position of the window
in one pass, so the window also holds the targets. One mean and one
standard deviation over the whole window would put information from
these targets into the inputs. We use an
exponentially weighted moving average (EWMA), whose statistics at
step $t$ read the values up to $t$:
\begin{align*}
  m_t &= (1 - \alpha)\, m_{t-1} + \alpha\, x_t, \\
  v_t &= (1 - \alpha)\, v_{t-1} + \alpha\, (x_t - m_t)^2, \\
  z_t &= \frac{x_t - m_t}{\max\big(\sqrt{v_t},\, 10^{-5}\big)}.
\end{align*}
We set $\alpha = 2 / (s + 1)$ with the span $s = 128$. This gives
$\alpha \approx 0.0155$ and a half-life of about $44$ values. The
recursion starts at $t = 0$ from $m_{-1}$ and $v_{-1}$, the mean and
the variance of the first $16$ values, so the statistics inside the
first patch read the whole patch. When a series starts with zero padding
(Section~\ref{sec:data}), the statistics start at the first value that
is not zero (a real zero at the start counts as padding) and the
padding stays at $0$. If the padding ends inside a patch, the first
$16$ real values extend into the next patch, and $m_{-1}$ and
$v_{-1}$ include values from the target of the patch that holds the
first real value. The losses therefore skip this patch.

At test time, the future values are unknown, so the statistics stop at
the last value of the context. The forecast $\hat{y}$ of every future
step goes back to values as $\hat{y}\,\sigma + \mu$, with $\mu = m_t$
and $\sigma = \max(\sqrt{v_t}, 10^{-5})$ at this last value.

\subsection{Embedding of the time scale}
\label{sec:scale-embedding}

The normalisation removes the mean and the standard deviation of the
series, and the model gets no input about them. The model reads the time scale
of the series: its sampling frequency and its seasonal period. The
frequency has $15$ classes:
\[
  \{4, 10\}\,\mathrm{s}, \quad
  \{1, 5, 10, 15, 30\}\,\mathrm{min}, \quad
  \{1, 6\}\,\mathrm{h}, \quad
  \{\text{day}, \text{week}, \text{month}, \text{quarter}, \text{year}\}, \quad
  \text{unknown}.
\]
The seasonal period $p$ is the number of steps in one season, as
GluonTS defines it for the frequency, for example $p = 24$ for hourly
data. The bucket of $p$ is its smallest upper bound in
\[
  \{1, 4, 8, 16, 32, 64, 128, 256, 512, \infty\},
\]
so there are $10$ buckets, and an unknown period goes to the bucket of
$1$. Each frequency class and each bucket has a
learned embedding of size $3$. We append the two embeddings to the $W$
values of each patch, so the patch embedding reads $W' = W + 6 = 22$
numbers.

\subsection{Patch embedding}
\label{sec:patch-embedding}

TimesFM~\citep{das2024timesfm} embeds a patch with the residual block of
TiDE~\citep{das2023tide}: a small MLP (linear, activation, linear) of
the patch values, summed with a linear map of the same values, then a
layer normalisation. The first layer of the MLP reads the patch as one
vector of $W$ values.

Our patch embedding is a bidirectional GRU. It reads the $W'$ numbers
of a patch as a sequence of $W'$ scalar steps: the $W$ values in time
order, then the $6$ numbers of the time-scale embeddings. This gives
the patch embedding a chance to model the local temporal structure
that an MLP collapses into a single linear projection. The GRU has
$2$ layers with $128$ units in each direction. A linear map takes the
final hidden states of the two directions ($256$ numbers) to
$\mathbb{R}^{\dmodel}$. We add a linear map of the $W'$ input numbers, and a
layer normalisation gives $e_t$.

\subsection{Encoder and forecaster}
\label{sec:forecaster}

\begin{figure}[t]
  \centering
  \includegraphics[width=\linewidth]{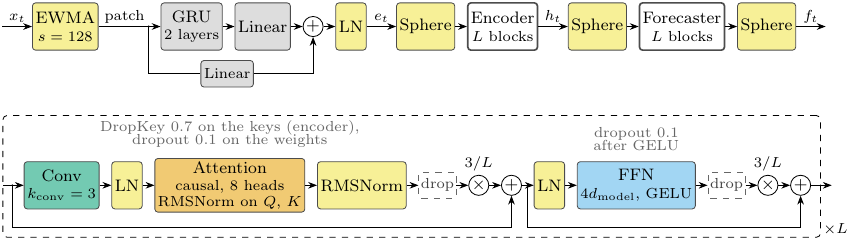}
  \caption{The model (top), and the block of the encoder and
    the forecaster (bottom). LN is a layer
    normalisation. A stack repeats the block $L$ times ($6$ in the
    encoder and $6$ in the forecaster), then puts its output on the sphere
    of radius $\sqrt{\dmodel}$. A dashed box marks a dropout of $0.1$.
    DropKey and dropout act only during training.}
  \label{fig:block}
\end{figure}

The encoder and the forecaster are causal transformers with the same
block (Figure~\ref{fig:block}), of width $\dmodel$ and with
$n_{\mathrm{head}} = 8$ attention heads. In each block, a depthwise causal convolution of kernel size
$k_{\mathrm{conv}} = 3$ reads the stream, a
layer normalisation follows, and the self-attention reads the result.
The residual connection adds the attention output to the stream as it
was before the convolution. An RMSNorm normalises the queries and the
keys of each head, and another one normalises the attention output. A
feed-forward network of hidden width $4\dmodel$ with a GELU follows, with a
pre-norm residual connection. The convolution follows the observation
of Elhage et al.~\citep{elhage2021circuits} that a short-range
depthwise convolution lets induction-head-like circuits form without
the second attention layer that the canonical induction circuit
requires. Our model has $6$ encoder layers and $6$ forecaster
layers (Section~\ref{sec:experiments}).
The initialisation follows muP~\citep{yang2022mup}. The base model has
width $64$ and $3$ layers in each stack. At our width $320$, the
muP rules scale the learning rate of the hidden weights by
$64 / 320$. Each weight starts at the PyTorch initialisation
times a factor that depends on its class:
\[
  \text{input}\ \times 0.3, \qquad
  \text{hidden}\ \times 1, \qquad
  \text{output}\ \times 0.
\]
The input weights are the projections of the patch embedding, and the
output weights write the output of the last layer of each stack. The
residual branches of a stack of $L$ layers are multiplied by $3 / L$,
so by $1 / 2$ in each of our stacks of $6$ layers.

\paragraph{DropKey and dropout.}
During training, DropKey~\citep{li2023dropkey} hides past patches from
the attention of the encoder: for each patch $t$ and each past patch
$j < t$, patch $t$ cannot attend to patch $j$ with probability $0.7$.
One random mask serves each series, for all heads and layers, and a
patch always attends to itself. DropKey removes keys before the
softmax, so the weights of the other keys still sum to $1$. The teacher
hides no patch (Section~\ref{sec:loss}), so the query and the positive
key of the loss term $\mathcal{L}_{\mathrm{rep}}$ read two different views of the
same past. Both
stacks also use a standard dropout of $0.1$ in each block: on the
attention weights, on the attention output, in the hidden layer of the
feed-forward network, and on the output of the feed-forward network.

\subsection{Loss function}
\label{sec:loss}

The targets of the loss come from a teacher: a copy of the patch
embedding and of the encoder, whose weights are an exponential moving
average (EMA) of the weights of the model. The teacher shares the live
time-scale embeddings of the model, with no EMA. After each step, the
teacher weights become $\theta_T \leftarrow \alpha\,\theta_T +
(1 - \alpha)\,\theta$, with a linear schedule of $\alpha$ from
$0.99995$ at the first step to $1$ at the last step. The teacher reads the full context, with no hidden
patch and no dropout, and no gradient flows through it. We write $h^T_t$ for its
encoder latent of patch $t$. The loss has four terms:
\begin{equation}
  \mathcal{L} \;=\;
  \mathrm{SIGReg}(e) \;+\; \mathrm{SIGReg}(h)
  \;+\; 0.1\, \mathcal{L}_{\mathrm{rep}}
  \;+\; 10\, \mathcal{L}_{\mathrm{align}}.
  \label{eq:loss-total}
\end{equation}
Every term skips the zero padding.

\paragraph{SIGReg on the embedding and on the encoder latent.}
A loss between two latents has a collapsed minimum: the same latent for
every patch. SIGReg~\citep{balestriero2025lejepa} pushes the latents to
an isotropic Gaussian distribution, and a collapsed distribution is far
from it. SIGReg projects the latents of the batch on random unit
directions and compares the distribution of each projection with a
Gaussian through their characteristic functions. For the latents
$z_1, \ldots, z_N \in \mathbb{R}^{\dmodel}$ of all the real patches of the
batch, with the anchor skips of Section~\ref{sec:normalisation}, and $M = 1024$ unit directions $a_1, \ldots, a_M$ drawn again
at each step,
\begin{equation*}
  \mathrm{SIGReg}(z) \;=\; \frac{1}{M} \sum_{m=1}^{M}
  \int_{-6\sqrt{\dmodel}}^{6\sqrt{\dmodel}}
  \bigg|\, \frac{1}{N} \sum_{n=1}^{N}
  \exp\!\Big(i\,\omega\, \frac{\langle a_m, z_n \rangle}{\sqrt{\dmodel}}\Big)
  \;-\; \exp\!\Big(-\frac{\omega^2}{2\dmodel}\Big) \bigg|^2 \, d\omega .
\end{equation*}
The target of each projection of $z / \sqrt{\dmodel}$ is the Gaussian
$\mathcal{N}(0, 1/\dmodel)$, so each coordinate of $z$ gets a variance close
to $1$, the scale of the sphere. SIGReg reads the embeddings $e$ and
the encoder latents $h$ before their projection onto the sphere.

\paragraph{Contrastive loss on the representation.}
$\mathcal{L}_{\mathrm{rep}}$ is a contrastive loss on the encoder
latent, in the style of MoCo~\citep{he2020moco}. For the anchor
$(b, t)$, patch $t$ of series $b$, the query is the latent $h_{b,t}$
of the model, which reads a context with hidden patches. The positive
key is the teacher latent $h^T_{b,t}$ of the same patch, which reads
the full context. The negative keys are the teacher latents of the
other patches of series $b$ and of all the patches of the other series
of the batch. The loss is
\begin{equation*}
  \mathcal{L}_{\mathrm{rep}} \;=\; -\frac{1}{|\mathcal{A}|}
  \sum_{(b,t) \in \mathcal{A}}
  \log \frac{e^{\mathrm{sim}(h_{b,t},\, h^T_{b,t}) / \tau}}
  {e^{\mathrm{sim}(h_{b,t},\, h^T_{b,t}) / \tau}
   + \sum_{q,\,k} e^{\mathrm{sim}(q, k) / \tau}},
\end{equation*}
with $\mathrm{sim}$ the cosine similarity, the temperature $\tau = 1$,
and $\mathcal{A}$ the set of the anchors of the batch. The inner
sum runs over the queries $q$ of all the anchors at time $t$ and over
their negative keys $k$. All the anchors at time $t$ share this sum.
$\mathcal{L}_{\mathrm{rep}}$ makes the latent of a patch closer to the
teacher latent of the same patch than to the latents of the other
patches and of the other series. Every patch of the window is an
anchor, except the last one.

\paragraph{Alignment of the forecasts.}
$\mathcal{L}_{\mathrm{align}}$ ties each forecast to the teacher latent
of the patch that it predicts, at distances from $1$ to $4$ patches.
We write $F$ for the forecaster. The one-step forecasts are
$f^{(0)} = F(h)$, with $f^{(0)}_t = f_t$. For $d > 1$, the forecaster
reads its own forecasts in place of $h$, a latent rollout: $f^{(d-1)} = F(f^{(d-2)})$,
so $f^{(d-1)}_t$ is the forecast of patch $t + d$ from patch $t$. The
gradient flows through all the calls of $F$. The loss is
\begin{equation*}
  \mathcal{L}_{\mathrm{align}} \;=\; \sum_{d=1}^{4}\;
  \underset{t}{\mathrm{mean}}
  \Big[\, 2 - 2 \cos\!\big(f^{(d-1)}_t,\; \mathrm{sg}(h^T_{t+d})\big) \Big],
\end{equation*}
with $\mathrm{sg}$ the stop-gradient. The four distances have the same
weight.

On the sphere, this cosine loss is a mean squared error. For two
vectors $u, v$ of norm $\sqrt{\dmodel}$,
\[
  \frac{\|u - v\|^2}{\dmodel} \;=\; 2 - 2 \cos(u, v).
\]
The forecast lies on the sphere, and the cosine does not depend on the
norm of the teacher latent. So each term of
$\mathcal{L}_{\mathrm{align}}$ is the mean squared error per coordinate
between the forecast and the teacher latent put on the same sphere.

For the curious reader, the cosine loss also has a probabilistic
reading. The von Mises--Fisher distribution on the sphere, with mean
direction $\mu$ and concentration $\kappa > 0$, has a density
proportional to $\exp\big(\kappa \cos(x, \mu)\big)$. For a fixed
$\kappa$, and up to a scale and a constant, each term $2 - 2\cos(f, h^T)$ is thus the negative log-likelihood of
the teacher latent under a von Mises--Fisher distribution centred on
the forecast: $\mathcal{L}_{\mathrm{align}}$ is a cross-entropy on the
sphere. This is the next-token objective of a language model.
$\mathcal{L}_{\mathrm{align}}$ has no negatives: SIGReg and
$\mathcal{L}_{\mathrm{rep}}$ prevent the collapse.

\paragraph{Contrastive Forecasting: a single contrastive loss.}
\label{sec:joint-loss}
Our recipe trains with a contrastive loss on the representation,
$\mathcal{L}_{\mathrm{rep}}$, and an alignment loss on the forecasts,
$\mathcal{L}_{\mathrm{align}}$. As a variant of the recipe,
we also train our model with one single contrastive loss in their
place,
$\mathcal{L}_{\mathrm{contrastive}}$: one InfoNCE-style fraction for each forecast distance, with every
key and every target from the EMA teacher. For the anchor
$(b, t)$ and the distance $d$, the forecast pair ties
$f^{(d-1)}_{b,t}$ to $h^T_{b,t+d}$, the representation pair ties
$h_{b,t}$ to $h^T_{b,t}$, and both pairs sit in the numerator:
\begin{equation*}
  \mathcal{L}_{\mathrm{contrastive}} = \sum_{d=1}^{4}
  -\frac{1}{|\mathcal{A}_d|} \sum_{(b,t) \in \mathcal{A}_d}
  \log \frac{e^{\mathrm{sim}(f^{(d-1)}_{b,t},\, h^T_{b,t+d}) / \tau}
   + e^{\mathrm{sim}(h_{b,t},\, h^T_{b,t}) / \tau}}
  {e^{\mathrm{sim}(f^{(d-1)}_{b,t},\, h^T_{b,t+d}) / \tau}
   + e^{\mathrm{sim}(h_{b,t},\, h^T_{b,t}) / \tau}
   + \sum_{q,\,k} e^{\mathrm{sim}(q,\, k) / \tau}},
\end{equation*}
with the temperature $\tau = 0.3$ and $\mathcal{A}_d$ the
anchors with a real patch at distance $d$. The inner sum runs over
the queries of all the anchors at time $t$ and over their negative
keys: the keys of $\mathcal{L}_{\mathrm{rep}}$, the teacher latents
$h^T_{b',t+d}$ of the other series, and the forecast
$f^{(d-1)}_{b,t+1}$ of the next patch of the same series. All the
anchors at time $t$ share this sum, as in
$\mathcal{L}_{\mathrm{rep}}$. The loss keeps the two SIGReg
terms of Equation~\eqref{eq:loss-total} and replaces
$0.1\, \mathcal{L}_{\mathrm{rep}} + 10\, \mathcal{L}_{\mathrm{align}}$
with $\mathcal{L}_{\mathrm{contrastive}}$, at weight $1$.

\subsection{Forecast head}
\label{sec:forecast-head}

\begin{figure}[t]
  \centering
  \includegraphics[width=\linewidth]{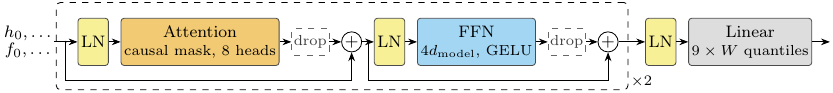}
  \caption{The forecast head. It reads the encoder latents and the
    forecasts of the frozen backbone, repeats a pre-norm block $2$
    times, and maps each token to $9$ quantiles of $W$ values. A dashed
    box marks a dropout of $0.1$, active only during training.}
  \label{fig:head-block}
\end{figure}

The forecast head maps the latents to values. We use it to compute all
our GIFT-Eval scores. The head is a standard pre-norm causal
transformer (Figure~\ref{fig:head-block}) with $2$ layers, width
$\dmodel$, $n_{\mathrm{head}} = 8$ attention heads and no positional
encoding. For each input token, a linear map gives $9$ quantiles, at
the levels $0.1, 0.2, \ldots, 0.9$, of the $W$ values of one patch. The
head has $2{,}512{,}784$ parameters.

\begin{figure}[t]
  \centering
  \includegraphics[width=0.88\linewidth]{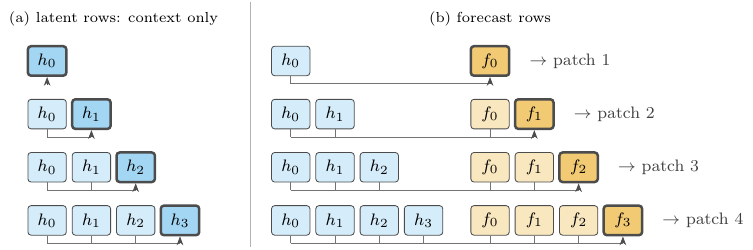}
  \caption{The attention of the forecast head during training, for
    $T_p = 4$. Each row shows the tokens that its bold token reads.
    (a) The latent $h_t$ reads $h_0, \ldots, h_t$. The head predicts
    nothing at $h_t$: the latents only give context to the forecasts.
    (b) The forecast $f_t$ reads $h_0, \ldots, h_t$ and
    $f_0, \ldots, f_t$, and its output predicts patch $t + 1$.}
  \label{fig:head-attention}
\end{figure}

\paragraph{Training.}
Once the backbone is trained, we freeze its weights and train the
forecast head on top of it. For a window of $T = 1024$ values, the frozen backbone gives the
encoder latents $h_0, \ldots, h_{T_p - 1}$ of the full context and the
one-step forecasts $f_0, \ldots, f_{T_p - 1}$. The latents are taken
before their projection onto the sphere, and the forecasts after it. The head reads a sequence
of $2 T_p$ tokens: the encoder latents, then the forecasts. The time
axis is repeated, so the forecast block starts again at patch $0$.
Figure~\ref{fig:head-attention} shows the tokens that each token
reads, for $T_p = 4$. The latent $h_t$ attends to $h_0, \ldots, h_t$.
The forecast $f_t$ attends to $h_0, \ldots, h_t$ and to
$f_0, \ldots, f_t$. The output at $f_t$ predicts patch $t + 1$. The latent
$h_{t+1}$ encodes this patch, so $f_t$ must not read it at any layer.
For this reason the latent rows are causal too: with a full latent
block, the first layer would carry $h_{t+1}$ into $h_t$, and $f_t$
would read it at the second layer. The loss is the pinball loss of the
$9$ quantiles against the values of patch $t + 1$, each value
normalised with the EWMA statistics of its own step. The loss is
averaged over the levels, the values and the forecast tokens, and the
outputs at the latent tokens get no loss. The padded targets get no
loss, and the head keeps the early anchors that the backbone losses
skip (Section~\ref{sec:normalisation}). A dropout of $0.1$ applies
in its blocks during training, at the same places as in the backbone
(Section~\ref{sec:forecaster}). Section~\ref{sec:experiments} gives
its optimiser settings. Appendix~\ref{app:head-no-attn} compares this head
with a head that reads one forecast latent and has no attention.

\begin{figure}[t]
  \centering
  \includegraphics[width=0.6\linewidth]{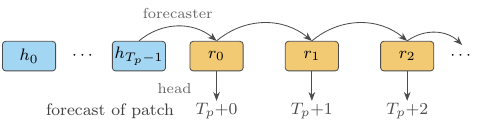}
  \caption{The forecast at test time. After the encoder latents of the
    context, the forecaster rolls out $r_0, r_1, \ldots$: each output
    is the next input. The head reads the whole sequence with a causal
    mask, and its output at $r_i$ is the forecast of patch $T_p + i$.}
  \label{fig:test-rollout}
\end{figure}

\paragraph{Forecast at test time.}
For a horizon of $H$ values, the head reads the context and a latent
rollout of the forecaster (Figure~\ref{fig:test-rollout}): the
forecaster runs on its own latent forecasts, with no value in between.
\begin{enumerate}
\item The backbone normalises the last $1024$ values of the context
  and encodes them into $h_0, \ldots, h_{T_p - 1}$. A shorter context
  gets zeros on its left.
\item The forecaster reads $h_0, \ldots, h_{T_p - 1}$, and its last
  output is $r_0$. For $i \geq 1$, it reads
  $h_0, \ldots, h_{T_p - 1}, r_0, \ldots, r_{i-1}$, each token on the
  sphere, and its last output is $r_i$. The rollout has
  $\lceil H / W \rceil$ steps. In the notation of
  $\mathcal{L}_{\mathrm{align}}$, $r_i = f^{(i)}_{T_p - 1}$.
\item The head reads $h_0, \ldots, h_{T_p - 1}, r_0, r_1, \ldots$ with a
  causal mask. Its output at $r_i$ gives the quantiles of patch
  $T_p + i$.
\item The EWMA statistics of the last value of the context map the
  quantiles back to values (Section~\ref{sec:normalisation}). The
  point forecast is the median.
\end{enumerate}

\section{Data}
\label{sec:data}

We train on the pretraining corpus of GIFT-Eval~\citep{aksu2024gifteval}
and we evaluate on the GIFT-Eval benchmark. The authors of GIFT-Eval
built this corpus with no overlap with the test split of the
benchmark. Each batch also holds a few synthetic series, and a mixup
step mixes the series of some batches.

\paragraph{Benchmark.}
GIFT-Eval has $23$ datasets with frequencies from $10$ seconds to one
year. They cover, for example, electricity load, solar power, weather,
road traffic, cloud servers, air quality, river flow, hospital
admissions, sales and the economic series of M4. Each dataset has up to
three forecast terms. The short term has a horizon set by the
frequency, for example $48$ steps for hourly data. The medium and long
terms have $10$ and $15$ times this horizon, and only $21$ pairs of a dataset
and a frequency, all hourly or finer, have them. A configuration is one dataset at one
frequency and one term, and the benchmark has $97$ configurations.

\paragraph{Real series.}
The pretraining corpus, GiftEvalPretrain, has $152$ sources,
$3.3$ million rows and $231$ billion values. We sample it with the
method of Moirai~\citep{woo2024moirai}. A source is drawn with a
probability proportional to its number of rows times the weight of the
source in the Moirai training mix. A row is drawn with a probability
proportional to its length (the ERA5 and CMIP6 sources draw their rows
uniformly, as in Moirai), and a window of $1024$ values starts at a
uniform position in the row. A draw keeps a random subset of the
variates of the row, as in Moirai, and each kept variate becomes a
separate univariate series. The covariates of a row train like its
target variates. A series shorter than the window gets zeros on its
left, and the normalisation of Section~\ref{sec:normalisation} treats
these zeros as padding. A run of $166{,}000$ steps reads $254$ real
windows at each step: $42$ million windows, or $43$ billion window
values. This is $18.7\%$ of the values of the corpus.

\begin{figure}[t]
  \centering
  \includegraphics[width=0.6\linewidth]{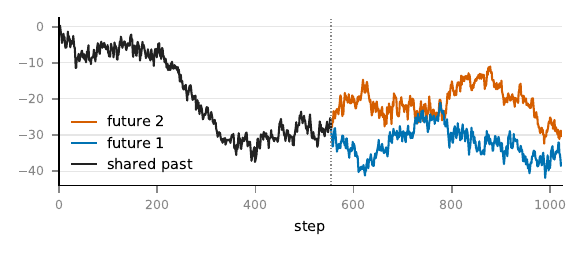}
  \caption{A forked ARMA pair: one shared past, then two futures after
    the fork (dotted line).}
  \label{fig:forked-arma}
\end{figure}

\paragraph{Synthetic series.}
Each batch holds two synthetic series, a \emph{forked ARMA} pair. The
two series share one past and have two different futures. We draw the
orders $p, q$ uniformly in $\{1, \ldots, 8\}$ and the coefficients
from a standard normal distribution, then we rescale each coefficient
set to an $\ell_1$ norm of at most $0.95$. This bound is a
sufficient condition for a stationary AR part and an invertible MA
part. The two series follow the same ARMA path
with unit Gaussian noise up to a fork point, drawn uniformly between
one quarter and three quarters of the window. After the fork, each
series continues from the shared state with its own noise, and with
its own copy of the coefficients plus a Gaussian perturbation of
standard deviation $0.05$. The perturbed coefficients are clipped to
the same $\ell_1$ bound, and the shared path starts after a burn-in of
$200$ steps. Each series is the cumulative sum of its
ARMA path, so it is an ARIMA$(p, 1, q)$ series.
Figure~\ref{fig:forked-arma} shows one pair.

\begin{figure}[t]
  \centering
  \includegraphics[width=0.6\linewidth]{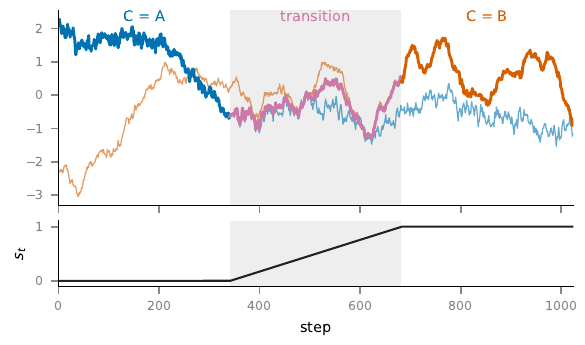}
  \caption{A crossfade $C$ (thick) of two series $A$ and $B$ (thin).
    The bottom panel gives the weight $s_t$. In
    training, $A$ and $B$ are real windows of the batch.}
  \label{fig:crossfade}
\end{figure}

\paragraph{Crossfade series.}
Each batch also holds three series built from two real windows $A$ and
$B$ of the batch: $A$ and $B$, each normalised to zero mean and unit
variance, and their crossfade
\[
  C_t \;=\; (1 - s_t)\, A_t \;+\; s_t\, B_t .
\]
The weight $s_t$ goes linearly from $0$ to $1$ over a transition
around a midpoint. The midpoint is uniform in the window, and the
width of the transition is log-uniform between $1/128$ of the window
and the full window (Figure~\ref{fig:crossfade}).

Each step has $259$ series: $254$ real windows, the forked ARMA
pair and the three crossfade series ($254 + 2 + 3 = 259$).

\paragraph{Mixup.}
At each step, with probability $0.3$, we mix the whole batch with the
mixup of~\citet{zhang2018mixup}. We draw one weight
$a \sim \mathrm{Beta}(0.2, 0.2)$ and one random permutation $\pi$ of
the batch, and each series $x_b$ becomes
\[
  \tilde{x}_b \;=\; a\, x_b \;+\; (1 - a)\, x_{\pi(b)} .
\]
The frequency and seasonality embeddings of
Section~\ref{sec:scale-embedding} mix with the same weight. The mixup
acts on the raw values, before the normalisation. A mixed series keeps
the longer zero padding of its two series. The TSMixup of
Chronos~\citep{ansari2024chronos} mixes up to $k$ series after a
scaling of each series; our mixup mixes two raw series. Before the
mixup, each series changes sign with probability $0.5$.

\section{Experiments}
\label{sec:experiments}

On GIFT-Eval, we compare our model with a value-space model
trained with the rest of our recipe.\footnote{Code, training scripts
and experiment notebooks are available at
\url{https://github.com/jeremycochoy/contrastive-forecasting}.}

\paragraph{Our model.}
Our model is the model of Section~\ref{sec:method}, trained with the
loss of Equation~\eqref{eq:loss-total} on the data of
Section~\ref{sec:data}. It has $\dmodel = 320$, $6$ encoder
layers and $6$ forecaster layers. The muP base stays at width $64$, so
the learning rate transfers. The residual scale $3 / L$ of
Section~\ref{sec:method} follows the depth rule of CompleteP with
base $3$, so the rate transfers across depth
too~\citep{dey2025completep}. It trains with AdamW ($\beta_1 = 0.9$,
$\beta_2 = 0.98$) and a weight decay of $0.1$, for the budget below.
The learning rate rises to $10^{-3}$ over the first $10{,}000$
steps, then stays constant to the end of training. The forecast head of Section~\ref{sec:forecast-head} then trains
on the frozen backbone for $30{,}000$ steps at batch size $256$, with
AdamW ($\beta_1 = 0.9$, $\beta_2 = 0.999$), a constant learning rate of
$10^{-3}$, a weight decay of $0.01$ and a gradient norm clipped at $1$.

\paragraph{The Moirai-style model.}
To measure the effect of the latent-space objective, we train a
value-space model with the rest of our recipe: the same data,
normalisation, time-scale embedding, GRU patch embedding and
transformer blocks. Its patches come from
Moirai~\citep{woo2024moirai}: it has one patch embedding and one
output head for each of the patch sizes $8$, $16$, $32$, $64$ and
$128$, and each series draws its patch size from the range of its
frequency.
Each output head predicts the $9$ quantiles of the next patch, and the
loss is the pinball loss on the normalised values. Like our model,
it runs on its own forecasts: the model reads its median
forecast up to $3$ times, and the loss adds the pinball loss of each
depth. The learning rate rises to $10^{-3}$ over $10{,}000$ steps,
then a cosine takes it to $0$ at step $166{,}000$. The model has
width $384$, $3$ encoder layers and $3$ forecaster layers: $14.4$
million parameters, the size of Moirai small. The baseline
differs from our run in a few more ways: it clips the gradient
norm at $1$, it skips the windows with NaN values or extreme
spikes, and it runs the attention, the feed-forward network and
the convolution in fp32. Our model runs them in bf16.

\paragraph{Training budget.}
Both models train for $166{,}000$ steps. They read $42$ million
windows of the pretraining corpus, about $20\%$ of its values.

\paragraph{Metric.}
For each of the $97$ configurations, we compute the MASE of the median
forecast, with the seasonal period of the frequency, and we divide it
by the MASE of the seasonal naive forecast. The GM-MASE is the
geometric mean of this ratio over the $97$ configurations. Lower is
better, and the seasonal naive forecast scores $1$. The GM-CRPS
repeats this with the CRPS of the nine quantile forecasts, divided by
the CRPS of seasonal naive, as on the GIFT-Eval leaderboard.
Each score comes from one training run, except the scores of our
model: they are the mean of two trainings (Section~\ref{sec:ablation}).

\section{Results}
\label{sec:results}

Table~\ref{tab:gift-eval} gives the GM-MASE and the GM-CRPS of
our model and of the Moirai-style value-space model of
Section~\ref{sec:experiments} after $166{,}000$
steps. Our model has width $320$, $6$ encoder and $6$ forecaster
layers, and a constant learning rate after the warmup. It is the
reference of
Section~\ref{sec:ablation}. Its scores are the mean of two
trainings with different seeds, $\pm$ half the noise of one training
(Section~\ref{sec:ablation}); the Moirai-style model has one
training. In GM-CRPS our model is clearly below the Moirai-style model. In
GM-MASE it is also below, but the gap is within the noise of one
training.

\begin{figure}[t]
  \centering
  \includegraphics[width=0.62\linewidth]{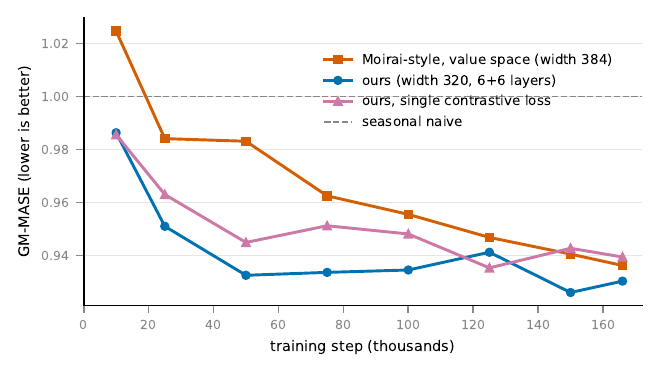}
  \caption{GM-MASE over the $97$ configurations at each scored
    stop: our model (width $320$, $6+6$ layers, constant learning
    rate), the Moirai-style model (width $384$, cosine schedule), and
    our model with the single contrastive loss.}
  \label{fig:gm-curves}
\end{figure}

\begin{figure}[t]
  \centering
  \includegraphics[width=0.62\linewidth]{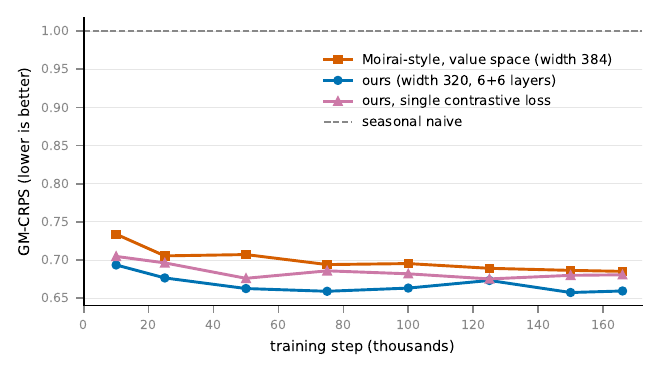}
  \caption{GM-CRPS over the $97$ configurations at each scored
    stop, for the runs of Figure~\ref{fig:gm-curves}.}
  \label{fig:gm-curves-crps}
\end{figure}

\begin{figure}[t]
  \centering
  \includegraphics[width=0.92\linewidth]{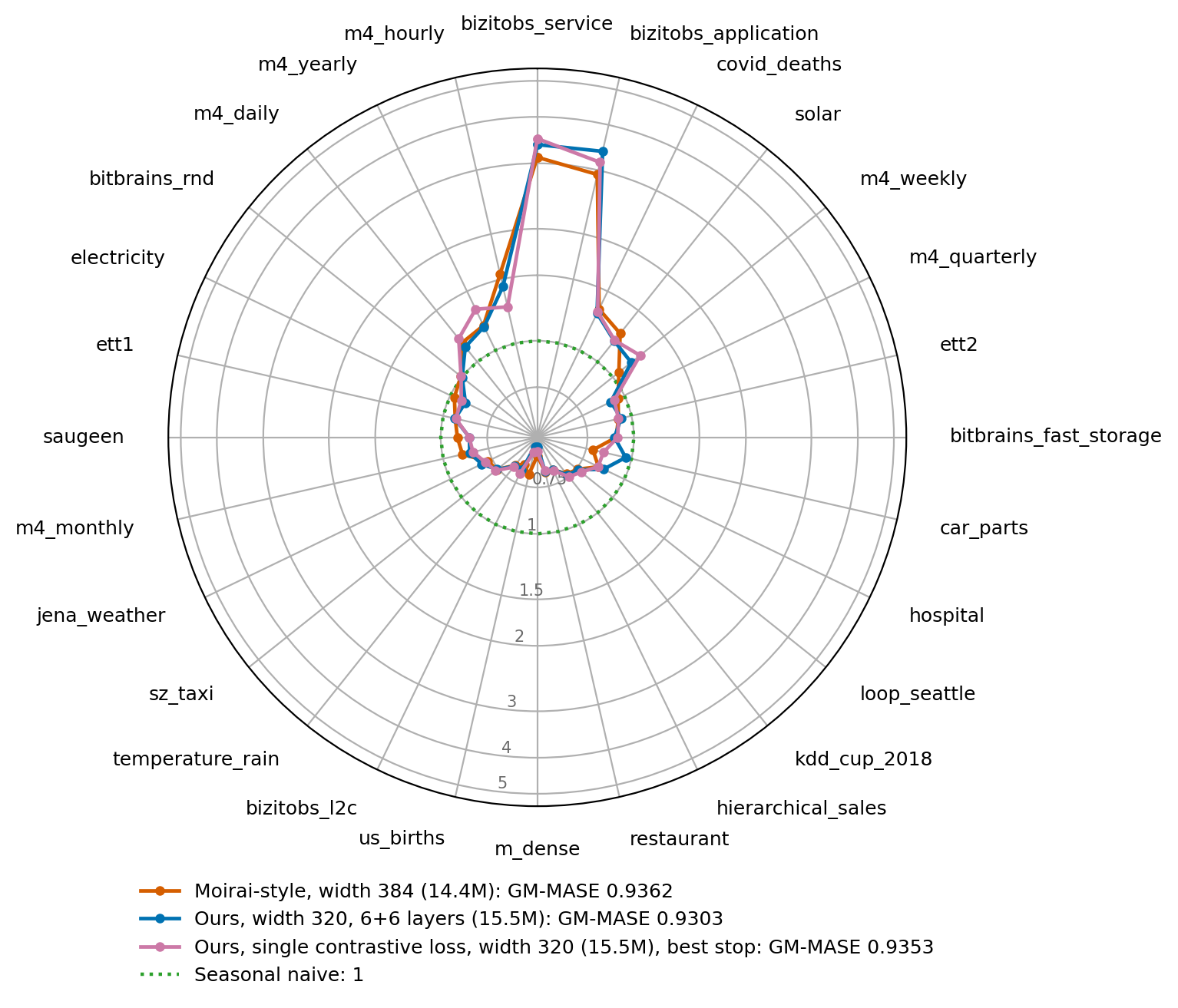}
  \caption{Relative MASE of each GIFT-Eval dataset, with M4 split by
    frequency: the geometric mean, over the
    configurations of the dataset,
    of the MASE of the model divided by the MASE of seasonal naive. The
    dotted ring is seasonal naive. Inside the ring is better. The
    hardest datasets are at the top. The legend gives the GM-MASE over
    the $97$ configurations. The purple line is our model trained
    with the single contrastive loss, at its best stop.}
  \label{fig:gift-radar}
\end{figure}

\begin{figure}[t]
  \centering
  \includegraphics[width=0.92\linewidth]{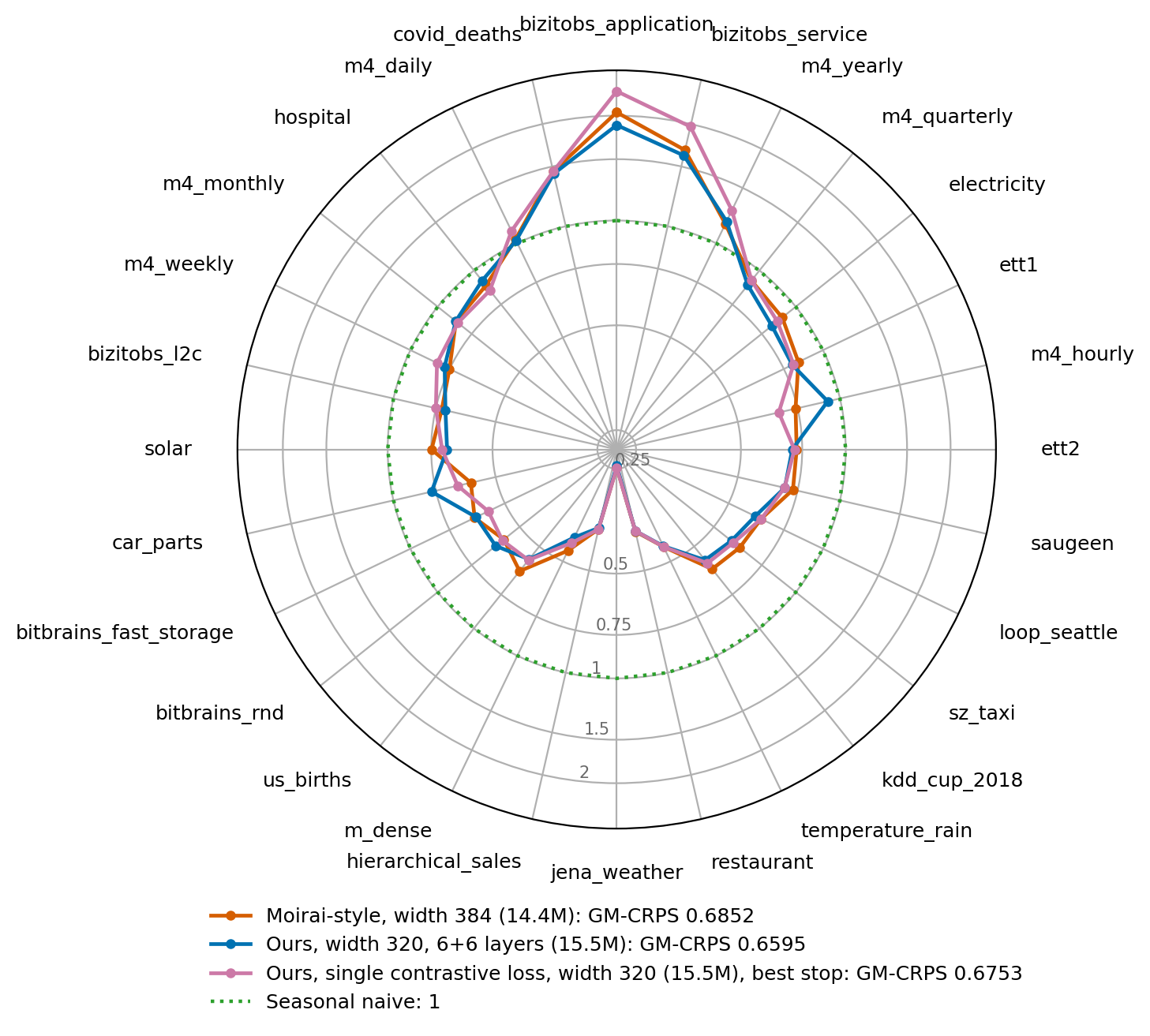}
  \caption{Relative CRPS of each GIFT-Eval dataset: the geometric
    mean, over the configurations of the dataset, of the CRPS of the
    model divided by the CRPS of seasonal naive. The models, the ring
    and the order of the datasets are those of
    Figure~\ref{fig:gift-radar}.}
  \label{fig:gift-radar-crps}
\end{figure}

Figure~\ref{fig:gm-curves} gives the GM-MASE along training. Our
learning rate is constant. The Moirai-style model follows a cosine
schedule to $0$. Our model is below it at every scored stop.
Figure~\ref{fig:gm-curves-crps} gives the same curves in
GM-CRPS. Both figures also show the run with the single
contrastive loss of Section~\ref{sec:joint-loss}.
Figure~\ref{fig:gift-radar} gives the
relative MASE of each GIFT-Eval dataset, and
Figure~\ref{fig:gift-radar-crps} the relative CRPS. Both radars
also show our model trained with the single contrastive loss of
Section~\ref{sec:joint-loss}
at its best stop. Appendix~\ref{app:horizon} gives the scores of these models
and of the ablations on each forecast horizon.

\begin{table}[t]
  \centering
  \caption{GM-MASE and GM-CRPS on the $97$ configurations of
    GIFT-Eval. Lower is better, and seasonal naive
    scores $1$. The layers are the encoder layers plus the forecaster
    layers. Our parameter counts leave out the forecast head and
    the EMA teacher. The Moirai-style count includes its output heads:
    they are part of the model. Appendix~\ref{app:parameters}
    gives the exact counts.}
  \label{tab:gift-eval}
  \setlength{\tabcolsep}{5pt}
  \begin{tabular}{lccccc}
    \toprule
    Model & Width & Layers & Parameters & GM-MASE & GM-CRPS \\
    \midrule
    Latent Forecasting (ours) & $320$ & $6 + 6$ & $15.5$M & $0.930 \pm 0.009$ & $0.663 \pm 0.007$ \\
    Moirai-style, value space & $384$ & $3 + 3$ & $14.4$M & $0.936$ & $0.685$ \\
    Ours, single contrastive loss & $320$ & $6 + 6$ & $15.5$M & $0.939$ & $0.681$ \\
    \bottomrule
  \end{tabular}

\end{table}

\subsection{The single contrastive loss}

We also trained our model with the single contrastive loss of
Section~\ref{sec:joint-loss}. This run reaches its best score at $125{,}000$ steps: a GM-MASE of
$0.935$ and a GM-CRPS of $0.675$, close to our model. Later in
training its score rises again, so Figures~\ref{fig:gift-radar}
and~\ref{fig:gift-radar-crps} show it at that stop.
The best stop is picked on the benchmark itself. We
understand the convergence of this loss less well than that of our
recipe, and it needs further study.

\section{Ablation}
\label{sec:ablation}

Our recipe changes several parts of a standard time-series foundation
model. Table~\ref{tab:ablation} lists these changes. Each ablation
starts from the reference, our model of
Section~\ref{sec:results}, and replaces one change with its
standard counterpart. The other settings stay the same, except
for the row marked $^{*}$. A new forecast head trains on each
backbone (Section~\ref{sec:forecast-head}).

\begin{figure}[!htb]
  \centering
  \includegraphics[width=\linewidth]{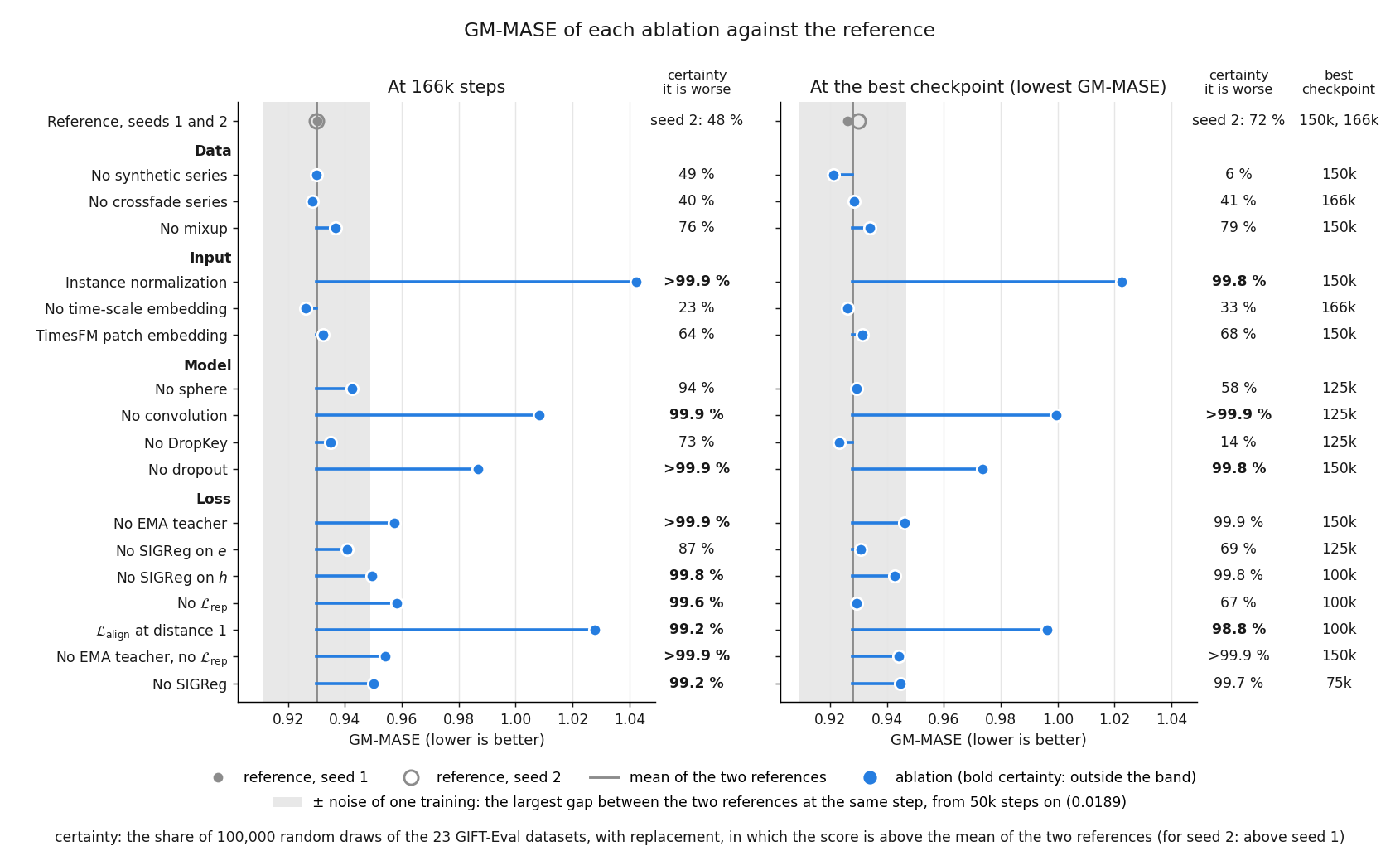}
  \caption{GM-MASE of each ablation at the end of training (left)
    and at its best stop (right), against the reference. The grey band
    is the noise of one training: the largest gap between the two
    trainings of the reference at the same stop. The certainty columns come from $100{,}000$ draws of the $23$
    GIFT-Eval datasets with replacement. They say how stable
    the gap is across the datasets, they do
    not measure the noise of training. Bold marks a score
    outside the band.}
  \label{fig:ablation-mase-best}
\end{figure}

\begin{figure}[!htb]
  \centering
  \includegraphics[width=\linewidth]{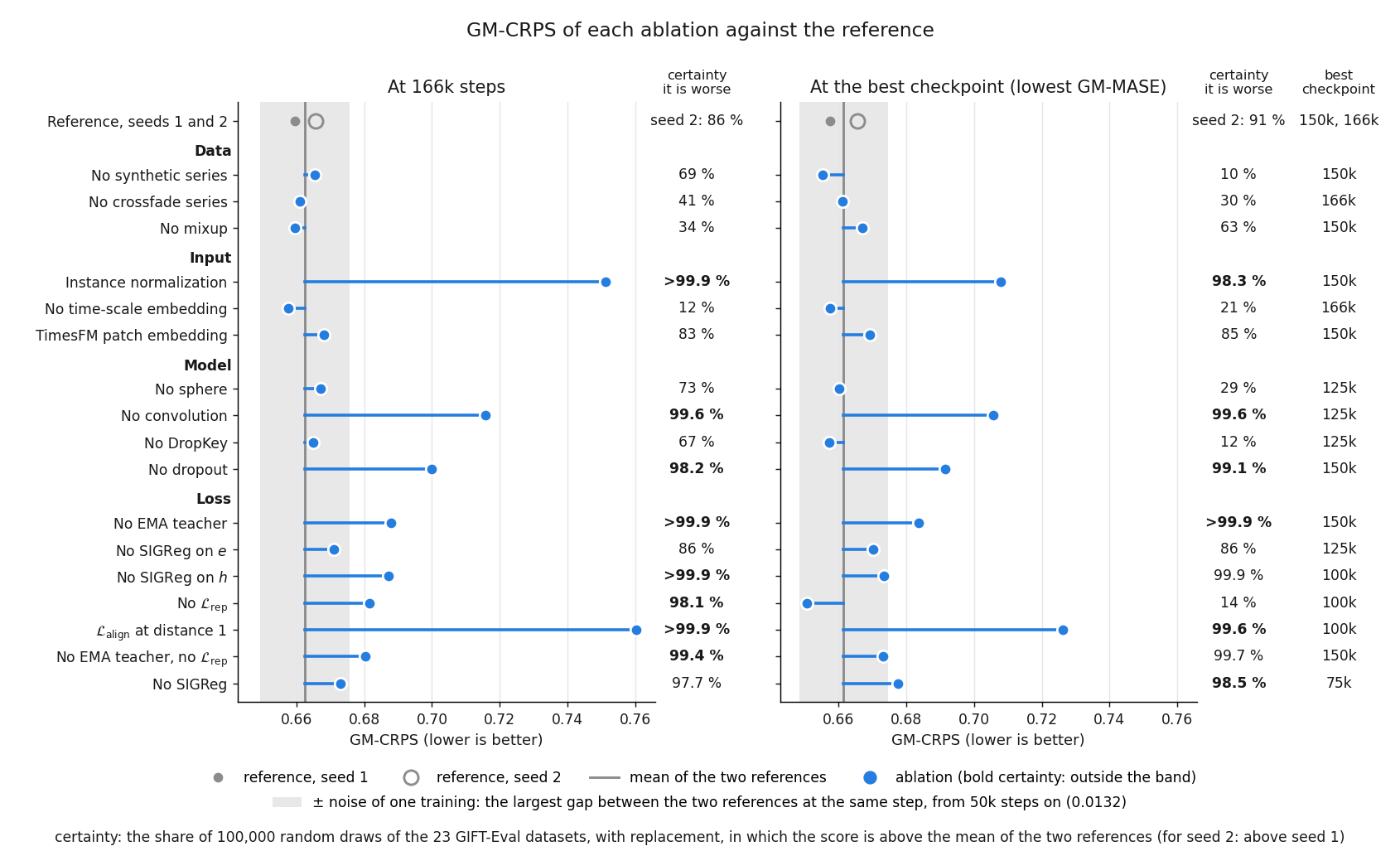}
  \caption{The GM-CRPS of the same runs, as in
    Figure~\ref{fig:ablation-mase-best}.}
  \label{fig:ablation-crps-best}
\end{figure}

To read the ablation scores better, we train the reference a
second time with another seed, despite our limited compute. The gap
between the two trainings gives an approximate idea of the training
noise. The reference row of Table~\ref{tab:ablation} is the mean
of the two, $\pm$ half this noise. At $166{,}000$ steps the two
trainings score $0.9303$ and $0.9299$ in GM-MASE, $0.6595$ and
$0.6656$ in GM-CRPS. Figures~\ref{fig:ablation-mase-best}
and~\ref{fig:ablation-crps-best} give each ablation at the
end of training and at its best stop. At the end of training, the
scores in bold in Table~\ref{tab:ablation} are worse than the
reference by more than this noise. Four variations stay worse at their
best stop too, so we can say that they degrade the
performance: instance normalisation, no convolution, no dropout, and
$\mathcal{L}_{\mathrm{align}}$ at distance $1$.

A few ablations need more details.

\paragraph{No time-scale embedding.}
The two embedding tables of Section~\ref{sec:scale-embedding}
stay at zero and do not train.

\paragraph{Residual block of TimesFM.}
The residual block of TimesFM v1~\citep{das2024timesfm}
replaces the GRU patch embedding and its layer normalisation. It
reads each patch, with its time-scale embedding, as one vector.

\paragraph{Student targets with stop-gradient.}
We remove the teacher. The targets of $\mathcal{L}_{\mathrm{align}}$
and the keys of $\mathcal{L}_{\mathrm{rep}}$ are the latents of the
student on the full context, with stop-gradient, as in
SimSiam~\citep{chen2021simsiam}.

\paragraph{Instance normalisation$^{*}$.}
Instance normalisation, as in RevIN~\citep{kim2022revin} and
Moirai~\citep{woo2024moirai}, normalises each window with one mean and
one standard deviation. Over the whole window, these two statistics
leak future values into the input
(Section~\ref{sec:normalisation}). So each series draws a context
of one patch to half of its window. The mean and the standard
deviation of the context normalise the whole window. The losses keep
only the anchors whose next patch comes after the context. The
forecast head of this ablation uses the uni2ts split: it forecasts the
last $15\%$ to $50\%$ of the window~\citep{woo2024moirai}.

\paragraph{Two parts at once.}
The last two rows of Table~\ref{tab:ablation} remove two
parts at once: the EMA teacher together with
$\mathcal{L}_{\mathrm{rep}}$, and SIGReg on both $e$ and $h$.

Appendix~\ref{app:ablation-curves} gives the GM-MASE and
the GM-CRPS of every ablation at each scored stop, with the two
trainings of the reference.

\begin{table}[!htb]
  \centering
  \caption{Ablations of our recipe. Each ablation replaces one part of
    the reference recipe with its standard counterpart. GM-MASE
    and GM-CRPS on the $97$ configurations of GIFT-Eval after
    $166{,}000$ steps (lower is better). The reference row is the
    mean of its two trainings, $\pm$ half the largest gap
    between them at the same stop.
    Bold marks a score worse than the reference by more
    than the noise of one training.}
  \label{tab:ablation}
  \footnotesize
  \setlength{\tabcolsep}{4pt}
  \resizebox{\linewidth}{!}{%
  \begin{tabular}{lllll}
    \toprule
    Part & Our recipe & Replaced by & GM-MASE & GM-CRPS \\
    \midrule
    & \multicolumn{2}{l}{Reference: width $320$, $6$ encoder and $6$ forecaster layers} & $0.930 \pm 0.009$ & $0.663 \pm 0.007$ \\
    \midrule
    Data & Forked ARMA series (Sec.~\ref{sec:data}) & No synthetic series & $0.930$ & $0.665$ \\
    Data & Crossfade series (Sec.~\ref{sec:data}) & No crossfade series & $0.928$ & $0.661$ \\
    Data & Mixup, $p = 0.3$ (Sec.~\ref{sec:data}) & No mixup & $0.937$ & $0.659$ \\
    \midrule
    Input & EWMA normalisation & Instance normalisation$^{*}$ & $\mathbf{1.042}$ & $\mathbf{0.751}$ \\
    Input & Time-scale embedding & No time-scale embedding & $0.926$ & $0.657$ \\
    Input & GRU patch embedding & Residual block of TimesFM & $0.932$ & $0.668$ \\
    \midrule
    Model & Latent sphere after each stage & No sphere & $0.943$ & $0.667$ \\
    Model & Depthwise causal convolution & No convolution & $\mathbf{1.008}$ & $\mathbf{0.716}$ \\
    Model & DropKey in the encoder, $p = 0.7$ & No DropKey & $0.935$ & $0.665$ \\
    Model & Dropout of $0.1$ in the blocks & No dropout & $\mathbf{0.987}$ & $\mathbf{0.700}$ \\
    \midrule
    Loss & EMA teacher & Student targets with stop-gradient & $\mathbf{0.957}$ & $\mathbf{0.688}$ \\
    Loss & SIGReg on $e$ & No SIGReg on $e$ & $0.941$ & $0.671$ \\
    Loss & SIGReg on $h$ & No SIGReg on $h$ & $\mathbf{0.949}$ & $\mathbf{0.687}$ \\
    Loss & $\mathcal{L}_{\mathrm{rep}}$ & No $\mathcal{L}_{\mathrm{rep}}$ & $\mathbf{0.958}$ & $\mathbf{0.682}$ \\
    Loss & $\mathcal{L}_{\mathrm{align}}$ at distances $1$ to $4$ & Distance $1$ only & $\mathbf{1.028}$ & $\mathbf{0.760}$ \\
    \midrule
    Loss & EMA teacher and $\mathcal{L}_{\mathrm{rep}}$ & Both removed at once & $\mathbf{0.954}$ & $\mathbf{0.680}$ \\
    Loss & SIGReg on $e$ and on $h$ & No SIGReg at all & $\mathbf{0.950}$ & $0.673$ \\
    \bottomrule
  \end{tabular}}
\end{table}

\FloatBarrier
\section{Conclusion}
\label{sec:conclusion}

We follow the next-token objective of language models: our model
predicts the next patch of a series. Following JEPA, it predicts the
latent of that patch rather than its raw values, so its target
can leave out the noise of the series. A contrastive loss with negatives
makes the latents of different patches and series stay apart, so the
latents keep as much information about the series as they can.
SIGReg keeps them from collapsing (Section~\ref{sec:method}).
The whole
model trains without labels. To show that its latent forecasts hold the
information of the future values, and that this information is easy to
read, we freeze the model and train only a small head that maps the
latent forecasts back to values.

On GIFT-Eval (Section~\ref{sec:results}), our model forecasts as well
as a value-space model trained with the same data for the same
number of training steps. The task is harder for it: its backbone never sees a
loss on the values, and only the head, trained afterwards, reads
the values out of the latent forecasts. We take this result as evidence
that forecasting in latent space is a viable way to train a time-series
foundation model. Section~\ref{sec:ablation} measures the
effect of each part of our recipe.

The evaluation is still narrow. A latent space that keeps the information
of the series should also serve other tasks, such as the
classification of series and anomaly detection. Each task needs only
a small head on the frozen model, so the same backbone can be scored
on all of them. We have also not tried to fine-tune the backbone on a task, which
may score better than a head on a frozen model.

One limit of the method is that its training loss does not tell how
good the model is. A low loss in latent space can still hide a poor
representation. To judge a model, we must train the head and score the
forecasts on a benchmark, which costs time at each save. A metric in
latent space that predicts the forecast error would make the iteration
loop much faster.

\section*{Acknowledgements}

We thank Redstone Solution OÜ for providing the compute of this
research.

Claude Code (Anthropic Claude Opus 4.7, Claude Opus 5.5 and Claude
Fable 5) assisted in drafting this manuscript and in running some of the
experiments, under the author's direction. The author takes full
responsibility for the contents of this paper.

\bibliographystyle{plainnat}
\bibliography{references}

\appendix
\section{Ablation curves}
\label{app:ablation-curves}

\begin{figure}[!htb]
  \centering
  \includegraphics[width=\linewidth]{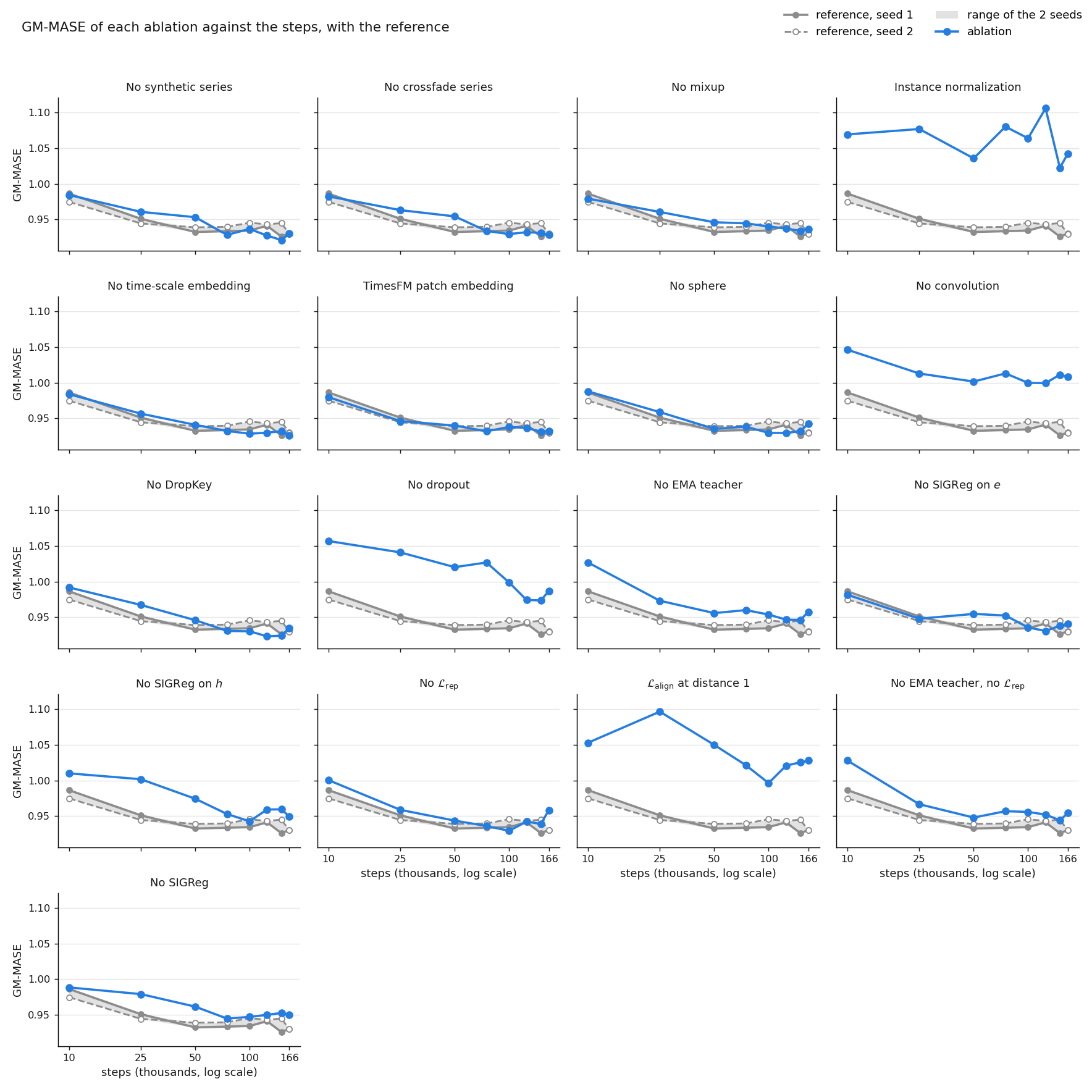}
  \caption{The GM-MASE of each ablation at each scored stop, on a
    log step axis. The grey lines are the two trainings of the
    reference, the shaded band is the range between them, and the blue
    line is the ablation.}
  \label{fig:ablation-mase-steps}
\end{figure}

\begin{figure}[!htb]
  \centering
  \includegraphics[width=\linewidth]{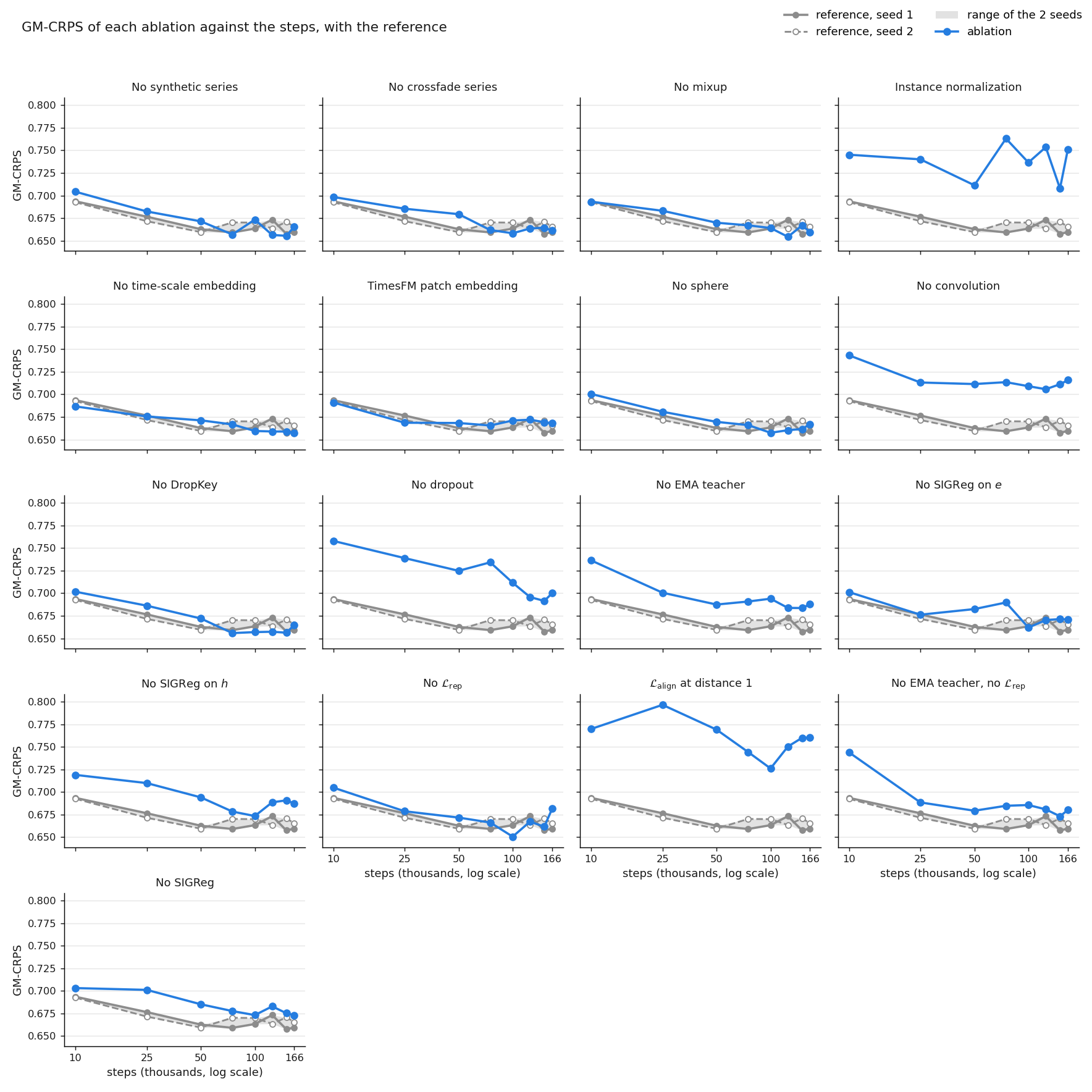}
  \caption{The GM-CRPS of the same runs at the same stops.}
  \label{fig:ablation-crps-steps}
\end{figure}

Figure~\ref{fig:ablation-mase-steps} gives the GM-MASE of
each ablation of Table~\ref{tab:ablation} at each scored stop,
and Figure~\ref{fig:ablation-crps-steps} the GM-CRPS. Each panel
shows one ablation against the two trainings of the reference, with
the range between them shaded.

\FloatBarrier
\section{A head without attention}
\label{app:head-no-attn}

The forecast head of Section~\ref{sec:forecast-head} reads
the latents of all the past patches. We also trained a head that
reads one forecast latent and nothing else. It is a stack of $8$
residual blocks of TimesFM~\citep{das2024timesfm}, with a
hidden width of $1280$ and a layer normalisation before each
block. The stack runs in the other direction: one latent gives the
$9$ quantiles of its patch. The head trains on the frozen backbone
with the data and the budget of the forecast head, and also on the
three rolled latents, which carry a quarter of the loss.

\begin{figure}[!htb]
  \centering
  \includegraphics[width=0.62\linewidth]{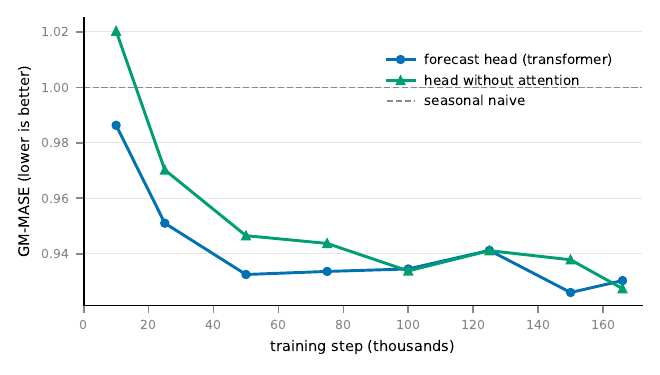}
  \caption{GM-MASE of our model at each scored stop, with the
    forecast head of Section~\ref{sec:forecast-head} and with
    the head without attention of Appendix~\ref{app:head-no-attn}.}
  \label{fig:gm-curves-head}
\end{figure}

Figure~\ref{fig:gm-curves-head} gives the GM-MASE of our
model at each scored stop, with each head. The two heads are close
from $100{,}000$ steps on, except at $150{,}000$. At the last stop
the head without attention scores $0.9274$, against $0.9303$ for the
forecast head. The gap is largest at the early stops.

\FloatBarrier
\section{Parameter counts and training compute}
\label{app:parameters}

The backbone of our model has $15{,}461{,}771$ parameters: the
patch embedding, the encoder and the forecaster, without the EMA
teacher. The forecast head has $2{,}512{,}784$ parameters. The
forecast at inference uses both: $17{,}974{,}555$ parameters. The
model with the single contrastive loss has the same architecture, so
the same counts. The Moirai-style model has $14.4$ million
parameters, with its five patch embeddings and its five output
heads. The output heads hold about $0.9$ million of them, so the
model has about $13.5$ million parameters without its heads.

The backbone trains at $1.9$ steps per second on one RTX 5090:
about $24$ hours for the $166{,}000$ steps. The forecast head trains
at $4.8$ steps per second on one RTX 4090: about two hours for its
$30{,}000$ steps. For the Moirai-style model, we use a second
training of the same model for $332{,}000$ steps, on one RTX 4080. It
ran at $0.8$ steps per second for about $115$ hours, so about $57$
hours for $166{,}000$ steps.

\FloatBarrier
\section{Scores by forecast horizon}
\label{app:horizon}

\begin{figure}[!htb]
  \centering
  \includegraphics[width=\linewidth]{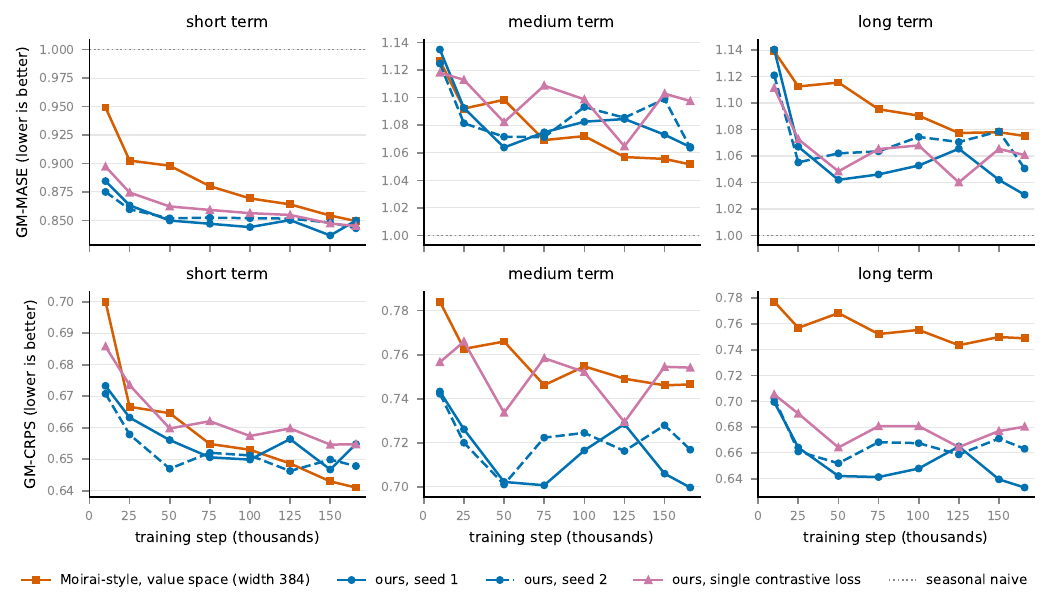}
  \caption{GM-MASE (top) and GM-CRPS (bottom) on each GIFT-Eval term
    at each scored stop: the two trainings of our model, our model with
    the single contrastive loss, and the Moirai-style model. The dotted
    line is seasonal naive.}
  \label{fig:horizon-models}
\end{figure}

\begin{figure}[!htb]
  \centering
  \includegraphics[width=\linewidth]{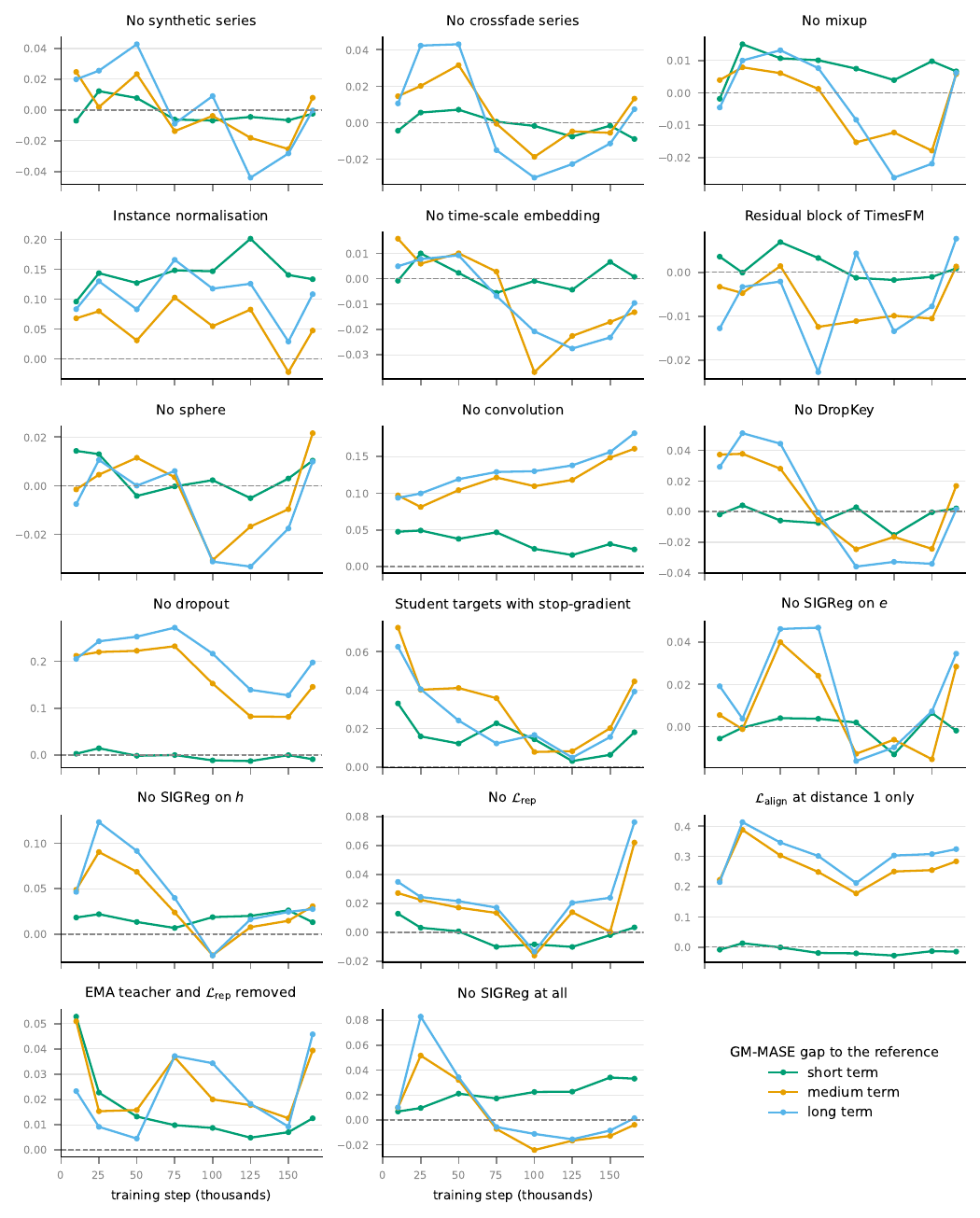}
  \caption{The GM-MASE of each ablation minus that of the reference, on
    each term, at each scored stop. The reference is the mean of its
    two trainings. Above the dashed line, the ablation is worse. Each
    panel has its own scale.}
  \label{fig:horizon-ablations-mase}
\end{figure}

\begin{figure}[!htb]
  \centering
  \includegraphics[width=\linewidth]{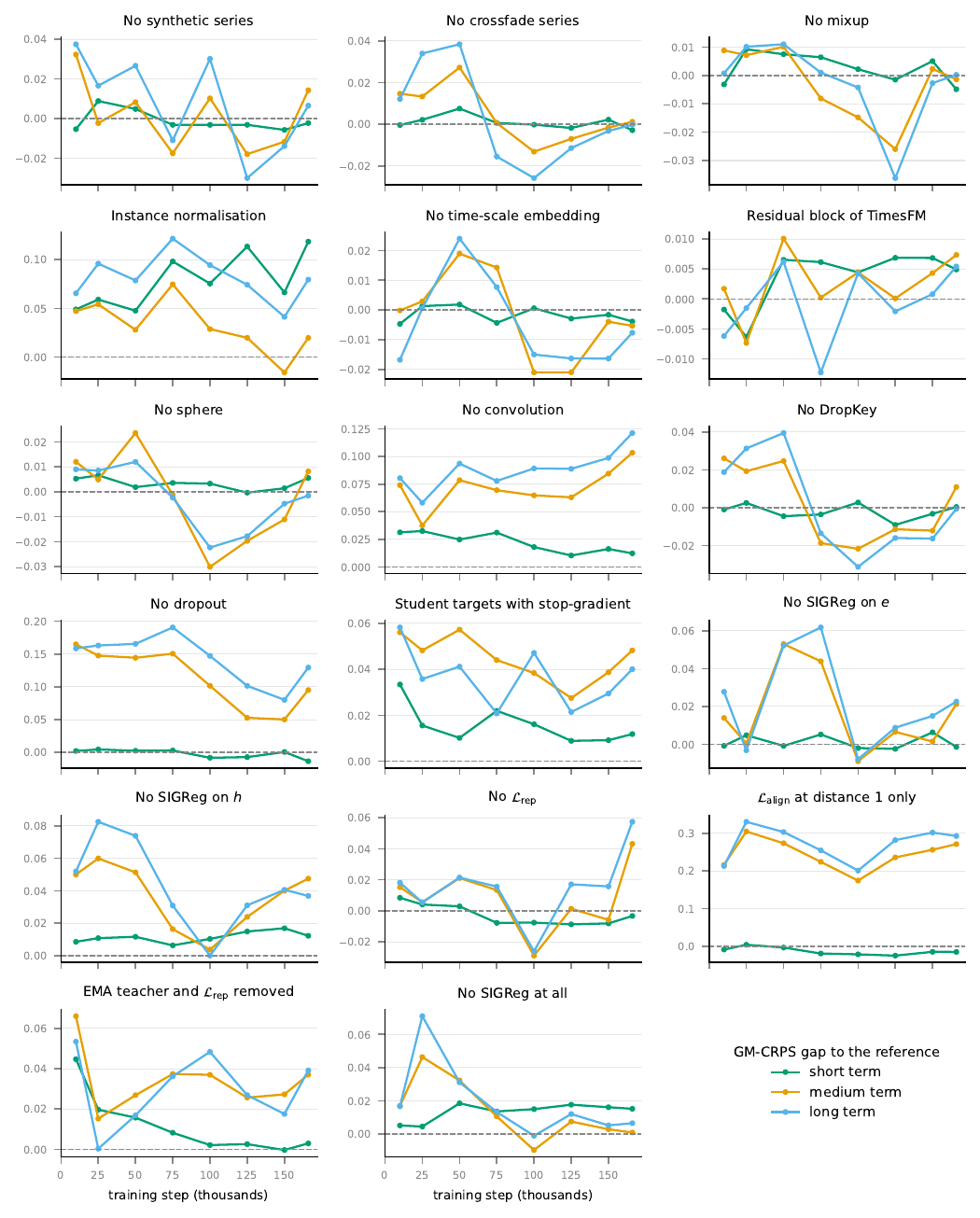}
  \caption{The same gaps in GM-CRPS.}
  \label{fig:horizon-ablations-crps}
\end{figure}

GIFT-Eval sorts its $97$ configurations into three terms: $55$
short-term, $21$ medium-term and $21$ long-term configurations. A
medium-term forecast has $10$ times as many steps as the short-term
forecast of the same dataset, and a long-term forecast $15$
times~\citep{aksu2024gifteval}. Table~\ref{tab:horizon} gives the
GM-MASE and the GM-CRPS of each term after $166{,}000$ steps, for the
models of Table~\ref{tab:gift-eval} and the ablations of
Table~\ref{tab:ablation}. Figure~\ref{fig:horizon-models} gives the
scores of the models on each term at each scored stop.
Figures~\ref{fig:horizon-ablations-mase}
and~\ref{fig:horizon-ablations-crps} give the gap between each
ablation and the reference on each term.

A term has fewer configurations than the full benchmark, so its
scores are noisier. On the long term, the two trainings of the
reference differ by $0.020$ in GM-MASE and by $0.030$ in GM-CRPS.

In GM-MASE, our model and the Moirai-style model score almost the
same on the short term. The Moirai-style model is lower on the medium
term, and our model on the long term. In GM-CRPS, the Moirai-style
model is lower on the short term, and our model on the medium and the
long terms.

Of the four ablations with the largest losses in
Table~\ref{tab:ablation}, three lose much more on the medium and the
long terms than on the short term: no convolution, no dropout and
$\mathcal{L}_{\mathrm{align}}$ at distance $1$ only. With no dropout
or with $\mathcal{L}_{\mathrm{align}}$ at distance $1$ only, the
short-term scores are even lower than those of both trainings of the
reference. The fourth, instance normalisation, loses most on the short
term.

\begin{table}[!htb]
  \centering
  \caption{GM-MASE and GM-CRPS on each GIFT-Eval term after $166{,}000$
    steps. Lower is better, and seasonal naive scores $1$. The short,
    medium and long terms have $55$, $21$ and $21$ configurations. The
    ablations are the rows of Table~\ref{tab:ablation}, in the same
    order.}
  \label{tab:horizon}
  \footnotesize
  \setlength{\tabcolsep}{4pt}
  \begin{tabular}{lcccccc}
    \toprule
    & \multicolumn{3}{c}{GM-MASE} & \multicolumn{3}{c}{GM-CRPS} \\
    \cmidrule(lr){2-4} \cmidrule(lr){5-7}
    Model & Short & Medium & Long & Short & Medium & Long \\
    \midrule
    Latent Forecasting (ours), seed $1$ & $0.850$ & $1.064$ & $1.031$ & $0.655$ & $0.700$ & $0.633$ \\
    Latent Forecasting (ours), seed $2$ & $0.843$ & $1.064$ & $1.051$ & $0.648$ & $0.717$ & $0.663$ \\
    Moirai-style, value space & $0.849$ & $1.052$ & $1.075$ & $0.641$ & $0.747$ & $0.749$ \\
    Ours, single contrastive loss & $0.845$ & $1.098$ & $1.061$ & $0.655$ & $0.754$ & $0.680$ \\
    \midrule
    No synthetic series & $0.844$ & $1.072$ & $1.040$ & $0.649$ & $0.723$ & $0.655$ \\
    No crossfade series & $0.838$ & $1.077$ & $1.048$ & $0.648$ & $0.710$ & $0.648$ \\
    No mixup & $0.853$ & $1.070$ & $1.047$ & $0.646$ & $0.707$ & $0.648$ \\
    \midrule
    Instance normalisation & $0.980$ & $1.111$ & $1.149$ & $0.770$ & $0.728$ & $0.728$ \\
    No time-scale embedding & $0.847$ & $1.051$ & $1.031$ & $0.647$ & $0.703$ & $0.641$ \\
    Residual block of TimesFM & $0.847$ & $1.065$ & $1.048$ & $0.656$ & $0.716$ & $0.654$ \\
    \midrule
    No sphere & $0.857$ & $1.085$ & $1.051$ & $0.657$ & $0.716$ & $0.647$ \\
    No convolution & $0.870$ & $1.224$ & $1.222$ & $0.664$ & $0.812$ & $0.769$ \\
    No DropKey & $0.849$ & $1.081$ & $1.042$ & $0.652$ & $0.719$ & $0.648$ \\
    No dropout & $0.837$ & $1.209$ & $1.238$ & $0.638$ & $0.803$ & $0.778$ \\
    \midrule
    Student targets with stop-gradient & $0.865$ & $1.108$ & $1.080$ & $0.663$ & $0.757$ & $0.688$ \\
    No SIGReg on $e$ & $0.845$ & $1.092$ & $1.075$ & $0.650$ & $0.730$ & $0.671$ \\
    No SIGReg on $h$ & $0.860$ & $1.095$ & $1.068$ & $0.664$ & $0.756$ & $0.685$ \\
    No $\mathcal{L}_{\mathrm{rep}}$ & $0.850$ & $1.126$ & $1.117$ & $0.648$ & $0.752$ & $0.706$ \\
    $\mathcal{L}_{\mathrm{align}}$ at distance $1$ only & $0.832$ & $1.348$ & $1.365$ & $0.636$ & $0.979$ & $0.941$ \\
    \midrule
    EMA teacher and $\mathcal{L}_{\mathrm{rep}}$ removed & $0.859$ & $1.103$ & $1.086$ & $0.654$ & $0.745$ & $0.687$ \\
    No SIGReg at all & $0.880$ & $1.060$ & $1.042$ & $0.667$ & $0.709$ & $0.655$ \\
    \bottomrule
  \end{tabular}
\end{table}

\end{document}